\documentclass[preprint,12pt, a4paper]{elsarticle}

\usepackage{amssymb}
\usepackage{hyperref}

\usepackage{soul,color}
\usepackage{listings}
\usepackage{xcolor}

\definecolor{codegreen}{rgb}{0,0.6,0}
\definecolor{codegray}{rgb}{0.5,0.5,0.5}
\definecolor{codepurple}{rgb}{0.58,0,0.82}
\definecolor{backcolour}{rgb}{0.95,0.95,0.92}

\lstdefinestyle{mystyle}{
    backgroundcolor=\color{backcolour},   
    commentstyle=\color{codegreen},
    keywordstyle=\color{magenta},
    numberstyle=\tiny\color{codegray},
    stringstyle=\color{codepurple},
    basicstyle=\ttfamily\footnotesize,
    breakatwhitespace=false,         
    breaklines=true,                 
    captionpos=b,                    
    keepspaces=true,                 
    numbers=left,                    
    numbersep=5pt,                  
    showspaces=false,                
    showstringspaces=false,
    showtabs=false,                  
    tabsize=2
}

\journal{Software Impacts}

\begin{document}

\begin{frontmatter}


\title{TopoDetect: Framework for Topological Features Detection in Graph Embeddings}

\author{Maroun Haddad and Mohamed Bouguessa}

\address{Department of Computer Science\\ University of Quebec at Montreal\\Montreal, Quebec, Canada\\haddad.maroun@uqam.ca, bouguessa.mohamed@uqam.ca}

\begin{abstract}
TopoDetect is a Python package that allows the user to investigate if important topological features, such as the Degree of the nodes, their Triangle Count, or their Local Clustering Score, are preserved in the embeddings of graph representation models. Additionally, the framework enables the visualization of the embeddings according to the distribution of the topological features among the nodes. Moreover, TopoDetect enables us to study the effect of the preservation of these features by evaluating the performance of the embeddings on downstream learning tasks such as clustering and classification.

\end{abstract}


\end{frontmatter}

\noindent

\section{Introduction}
In recent years, Graph representation models, in particular, Graph Neural Networks, have demonstrated state-of-the-art performance on numerous graph learning tasks \cite{Hamilton2017}. One of the main reasons for the success of these models is their capacity to generate vector representations (or embeddings) that capture important characteristics of the graph. However, little is known about the topological structures that are encoded in these embeddings and their effect on downstream learning. Therefore, we have designed \textit{TopoDetect} \cite{Haddad2021}, a framework that leverages well-established techniques in literature \cite{Rizi2017,Bonner2019} to determine whether certain topological features or centrality measures are encoded in the embeddings.

\section{TopoDetect cycle}
\begin{figure*}[!t]
\centerline{\includegraphics[scale=0.65]{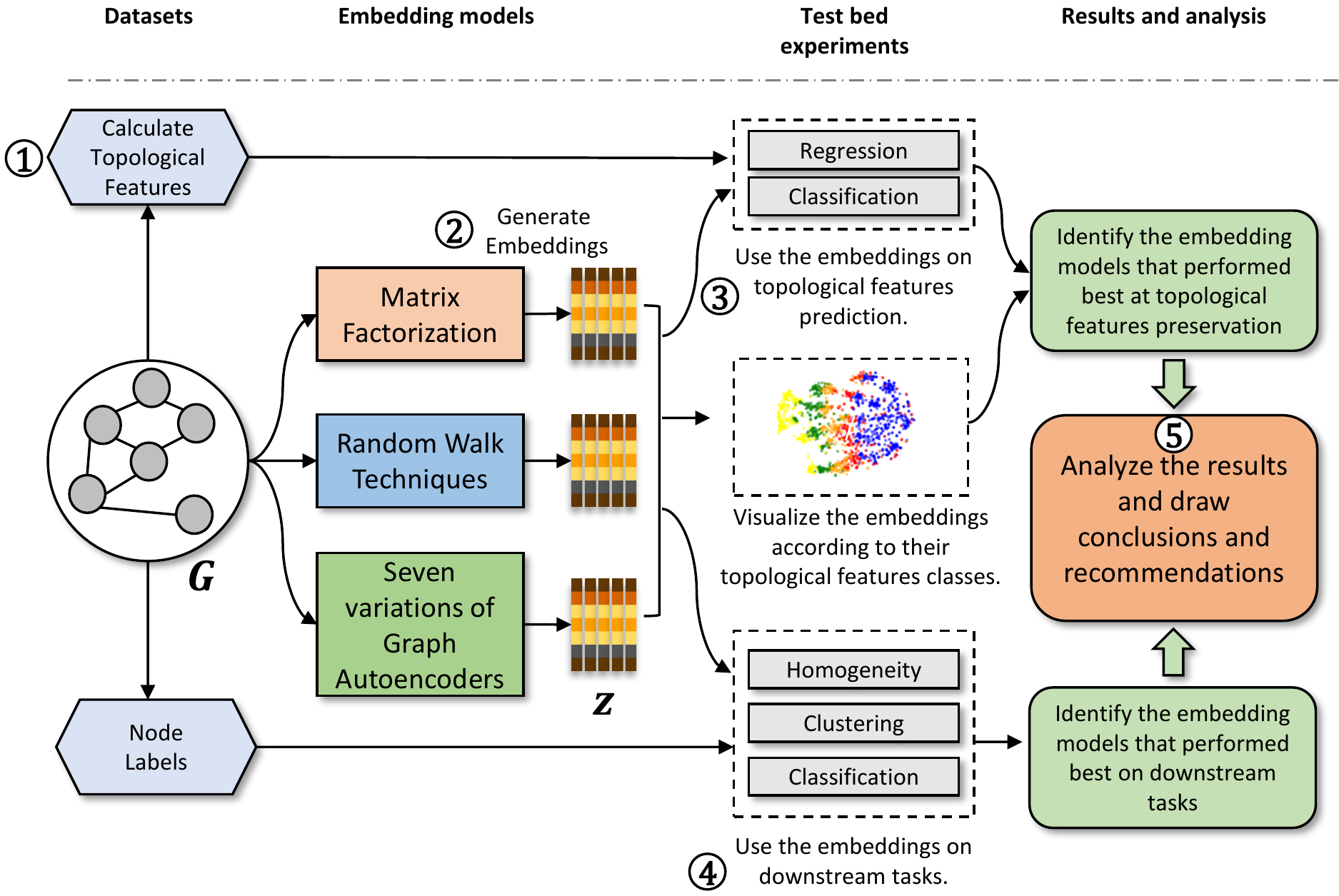}}
\caption{A flowchart of TopoDetect framework cycle.}
\label{fig:framework}
\end{figure*}

The TopoDetect framework is divided into five main steps, as illustrated in Figure \ref{fig:framework}. \textbf{In the first step}, we calculate the topological features of the nodes in the graph. \textbf{In the second step}, we use the graph representation learning models to generate embeddings for the nodes. \textbf{In the third step}, we investigate if the calculated topological features are encoded in the generated embeddings. For this step, two methods are utilized. In the first method, the topological features are directly predicted from the embeddings using Linear Regression \cite{Rizi2017}. The lower the prediction loss, the better the embeddings are at capturing the topological features. In the second method, we divide the topological features into different classes using histogram binning 
\cite{Bonner2019}. For example, all nodes whose Degree falls within a certain range, belong to the same class. Subsequently, we use classical classification models such as  Multilayer Perceptrons and Support Vector Machines to classify the nodes into their respective topological feature classes, using the embeddings as learning attributes. The better the model scores at the classification task, the better the embeddings are at capturing the topological features. \textbf{In the fourth step}, we use the embeddings on downstream learning tasks to cluster and classify the nodes according to their ground truth labels. \textbf{In the fifth and final step}, we analyze the results and conclude the effect of the preserved topological features on the downstream learning tasks.  

\section{Running experiments}
The framework includes by default ten datasets on which the user can directly run the experiments, such as Cora, Citeseer, and email-Eu-core. The datasets can be found under the folder \textbackslash data in the framework. To run an experiment, the user can simply call the \textit{run} function from the \textit{run\_experiments.py} file and pass the dataset name (as it appears in the \textbackslash data folder). Additionally, the user has to supply the number of bins to divide the topological feature classes and the number of runs for the experiments. The number of bins is determined experimentally, however, it is advised to use the number of bins that generates the most balanced classes between the features. The framework will display the mean and standard deviation of the results across all the runs. Below is an example on the Brazil Air-Traffic dataset.
 
\begin{lstlisting}[language=Python,frame=lines, caption=Predefined dataset experiment.]
import experiments.run_experiments as re

dataset = "brazil_airtraffic"
bins = 3
runs = 10

re.run(dataset, bins, runs)
\end{lstlisting}

The experiments can also be customized by selecting only certain models or downstream tasks for the tests. Furthermore, custom datasets can be used by calling the \textit{load\_custom\_dataset} from the \textit{utils.py} file.

\section{Visualization}
\begin{figure*}[!h]
\centerline{\includegraphics[scale=0.4]{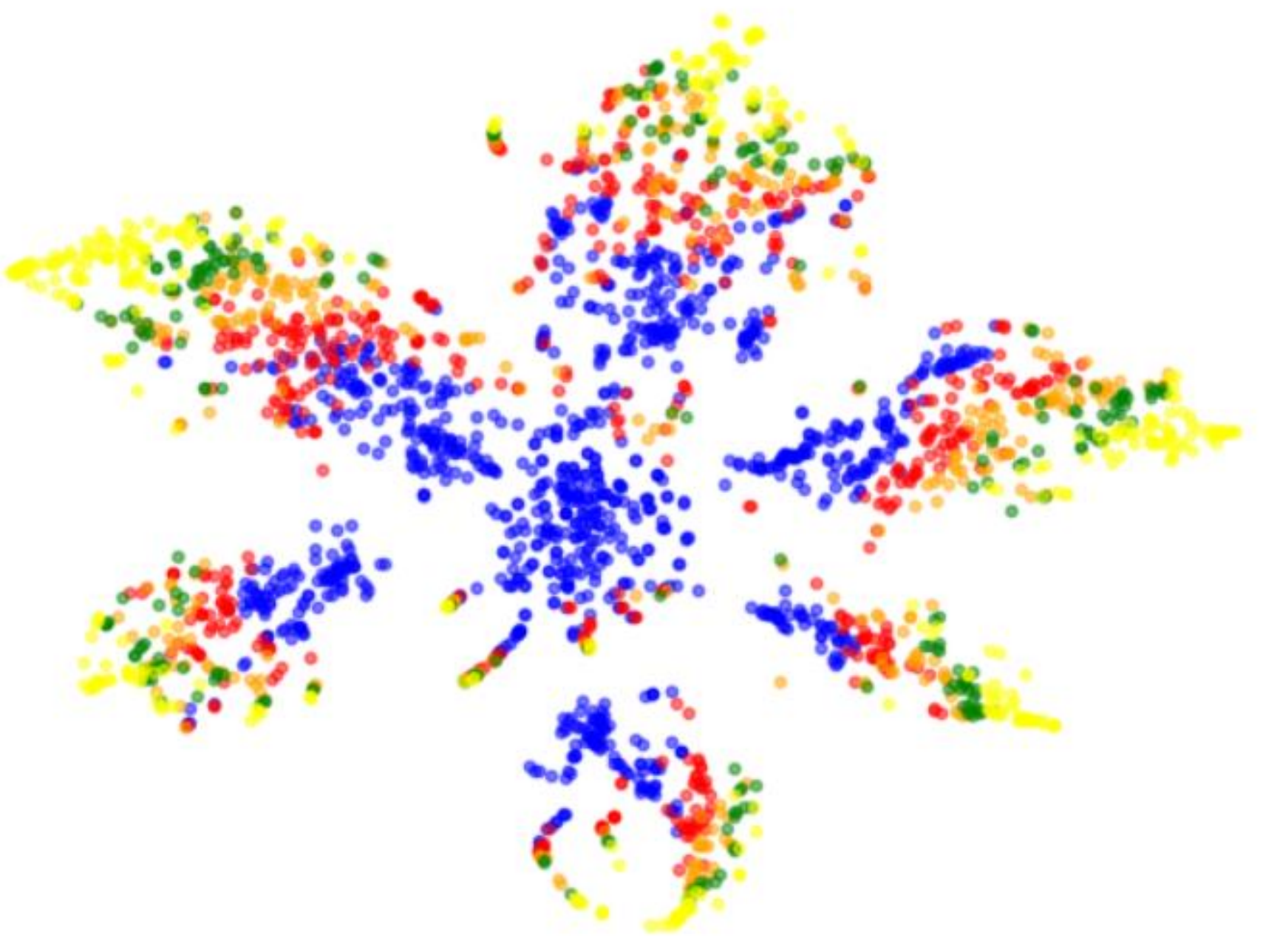}}
\caption{Visualization of the Cora dataset embedding generated with a Graph Autoencoder that uses the SUM aggregation rule and the first layer output as embedding. The nodes are colored according to their Degree class.}
\label{fig:embedding}
\end{figure*}

TopoDetect also enables the user to visualize the embeddings according to the topological classes generated by histogram binning. The embeddings are first projected into 2 dimensions using t-SNE \cite{Maaten2008}. Figure \ref{fig:embedding} shows the example of an embedding generated by a Graph Autoencoder that uses the SUM aggregation rule and the first layer output as embedding. Each Degree range (or class) was colored in a unique color. In this case, we can clearly observe that the nodes that belong to the same Degree class are neatly clustered together in the 2D projection. Such visualizations can help us confirm the preservation of the topological features in the embeddings of certain models. 
\section{Impact}
The TopoDetect framework has already permitted us to identify two variations of the Graph Autoencoder capable of capturing the topological features in their embeddings. These variations use the SUM rule for aggregating the messages, and the output of the first layer of the encoder in the embedding \cite{Haddad2021}. Our study showed that such models performed well on certain downstream tasks, where the preserved topological features are relevant to the learning task. Our findings were further validated in a case study, where the models that captured the Degree, Eigenvector, and Betweenness Centrality showed promising results on the task of social influence prediction. 

\section{Possible extensions}
By default, TopoDetect includes the exploration of five well-known topological features, which are: Degree, Triangle Count, Local Clustering Score, Eigenvector Centrality, and Betweenness Centrality. However, TopoDetect can be easily extended to investigate other topological features, for example, Katz Centrality or Page Rank. Additionally, in our study, we examined the embeddings of three categories of models: Matrix Factorization \cite{Belkin2001}, Random Walk \cite{Grover2016}, and Graph Autoencoders \cite{Kipf2016}. However, our framework can be utilized to test other types of models, most interestingly the Graph Attention Networks \cite{Velickovic2018}. Furthermore, TopoDetect can be extended to cover downstream learning tasks that go beyond clustering and classification. For example, users can study the effect of the preservation of topological features on the tasks of link prediction or edge labeling. Finally, even though our framework was used only in the context of directed and attributed graphs, it would be highly interesting to deploy it on other families of graphs, such as weighted graphs, multidimensional graphs, and even dynamic/temporal graphs. 

\bibliographystyle{elsarticle-num} 
\bibliography{References.bib}

\section*{Required Metadata}

\section*{Current code version}

Ancillary data table required for subversion of the codebase. Kindly replace examples in right column with the correct information about your current code, and leave the left column as it is.

\begin{table}[!ht]
\begin{tabular}{|l|p{6.5cm}|p{6.5cm}|}
\hline
\textbf{Nr.} & \textbf{Code metadata description} & \textbf{Please fill in this column} \\
\hline
C1 & Current code version & v1.0 \\
\hline
C2 & Permanent link to code/repository used for this code version &$https://github.com/MH-0/RPGAE$ \\
\hline
C3  & Permanent link to Reproducible Capsule & $https://codeocean.com/capsule/8314979/tree/v1$\\
\hline
C4 & Legal Code License   & MIT license \\
\hline
C5 & Code versioning system used & none \\
\hline
C6 & Software code languages, tools, and services used & python \\
\hline
C7 & Compilation requirements, operating environments \& dependencies & python 3.6, dgl-cu90 0.4.3, torch 1.4, networkx 2.4, numpy 1.18.1, scikit-learn 0.22.2, node2vec 0.3.2, cuda 9.0\\
\hline
C8 & If available Link to developer documentation/manual & $https://github.com/MH-0/RPGAE/blob/master/README.md$ \\
\hline
C9 & Support email for questions & haddad.maroun@uqam.ca, bouguessa.mohamed@uqam.ca\\
\hline
\end{tabular}
\caption{Code metadata.}
\end{table}

\end{document}